\documentclass{edm_article}
\usepackage{hyperref}
\usepackage{graphicx}
\usepackage{amsmath}
\usepackage{xcolor}
\usepackage{amssymb}
\usepackage{comment}

\newcommand{\hunter}[1]{\textcolor{teal}{\bf [Hunter: #1]}}

\begin{document}

\title{A Conceptual Model for End-to-End Causal Discovery in Knowledge Tracing}

\numberofauthors{1}

\author{
\alignauthor
{Nischal Ashok Kumar, Wanyong Feng, Jaewook Lee, Hunter McNichols, \\ Aritra Ghosh, Andrew Lan} \\
       \affaddr{University of Massachusetts Amherst}\\
    \email{{\small\{nashokkumar,wanyongfeng,jaewooklee,wmcnichols,arighosh,andrewlan\}@umass.edu}}
}

\maketitle

\begin{abstract}

In this paper, we take a preliminary step towards solving the problem of causal discovery in knowledge tracing, i.e., finding the underlying causal relationship among different skills from real-world student response data. This problem is important since it can potentially help us understand the causal relationship between different skills without extensive A/B testing, which can potentially help educators to design better curricula according to skill prerequisite information. Specifically, we propose a conceptual solution, a novel causal gated recurrent unit (GRU) module in a modified deep knowledge tracing model, which uses i) a learnable permutation matrix for causal ordering among skills and ii) an optionally learnable lower-triangular matrix for causal structure among skills. We also detail how to learn the model parameters in an end-to-end, differentiable way. Our solution placed among the top entries in Task 3 of the NeurIPS 2022 Challenge on Causal Insights for Learning Paths in Education. We detail preliminary experiments as evaluated on the challenge's public leaderboard since the ground truth causal structure has not been publicly released, making detailed local evaluation impossible. 

\end{abstract}



\section{Introduction}

Knowledge Tracing (KT) \cite{kt} refers to the problem of estimating a student's understanding or mastery of certain skills, concepts, or knowledge components through their responses to questions and using these estimates to predict future performance. KT methods are frequently utilized in modern online education platforms to determine the knowledge levels of many students to enable the platform to provide personalized feedback and recommendations, ultimately leading to better learning results \cite{ritter}. KT methods are limited in how they represent the relationship between skills; One key limitation is that most do not model the \textbf{causal} relationships between skills. Most KT methods simply treat human expert-provided skill tags as a flat structure (with a few exceptions, such as \cite{gikt}, that organize skills hierarchically as trees). As a result, these models are not capable of providing meaningful pedagogical insights, i.e., predicting future student performance if a particular instructional plan is applied instead of the actual plan applied. 

Causal analysis tools are a perfect fit to address these limitations in KT. The task of \emph{causal discovery}, i.e., learning causal relationships among different skills from observational data, is especially important. First, it helps educators learn prerequisite relationships among skills. This can guide educators in ordering topics within their curriculum and can guide students to review prerequisite information when they are stuck on a question \cite{desmarais2012mapping}. Second, causal relationships among skills helps us with the task of \emph{causal inference}, i.e., estimating the effect of a particular pedagogical treatment or intervention. Traditionally, these tasks are addressed through randomized controlled trials which are difficult to scale. Therefore, incorporating causal discovery into KT methods has the potential to become a scalable alternative since it can be done solely from observational student response data. Performing causal discovery directly from observational student response data is challenging since it is not straightforward to estimate treatment effects from observational data with incomplete or no knowledge of the causal relationship between skills. This problem is referred to as the \emph{end-to-end causal inference} problem, where we discover the causal graph and estimate treatment effects together. 

\subsection{Contributions}

In this paper, we take a \textbf{preliminary} step towards learning causal ordering among skills from student response data. This task is proposed in the NeurIPS 2022 Challenge on Causal Insights for Learning Paths in Education\footnote{\url{https://eedi.com/projects/neurips-2022}}. 
Our proposed \textbf{conceptual} solution is, to the best of our knowledge, the first KT method to learn the causal structure among human expert-provided skill tags directly from observational data in an \emph{end-to-end} manner. Specifically, our contributions in this paper are as follows:
\sloppy
\begin{itemize}
    \item First, we propose an interpretable \emph{causal structure} model that characterizes both i) the dependency among skills using a lower-triangular matrix and ii) their prerequisite ordering using a permutation matrix. We hypothesize that this module can be combined with any existing KT method that rely on human expert-provided skill tags. 
    \item Second, as a (among the top) solution\footnote{The code for our solution can be found at: \url{https://github.com/umass-ml4ed/Neurips-Challenge-22}} to Task 3 in the NeurIPS 2022 Challenge, we apply our causal structure module to a variant of deep knowledge tracing (DKT) \cite{dkt}, with a \emph{causal} gated recurrent unit (GRU) module at its core, due to i) the simple nature of DKT and ii) its good empirical performance in our experiments.
    \item Third, we detail our experimental results based on the public leaderboard of the NeurIPS 2022 Challenge. We are honest up front that our evaluation is limited since i) the ground-truth causal structure data is not publicly released and ii) the nature of this brand new task means that there are no baselines to compare against. 
\end{itemize}


\section{Related Work}

\subsection{KT methods}
\label{sec:rwkt}
Existing KT methods can be classified along several different axes, the first of which is how they represent the student knowledge representation variable $h$. Classic Bayesian KT methods, such as those in \cite{mozerfuse,ktcomparepardos,yudelson}, treat student knowledge as a latent binary variable. 
Recent methods like deep learning-based KT methods, such as \cite{akt,rkt,dkt,saint+,dkvmn}, treat student knowledge as hidden states in neural networks. This setup results in models that excel at predicting future performance but have limited interpretability \cite{kenjedm}. Another major axis is how KT methods represent responses, questions, skills, and time steps. To represent student responses, most existing KT methods treat them as binary-valued indicating response correctness. However, a few methods, such as option tracing \cite{ot} and predict partial analysis \cite{neilkt}, have characterized student responses as non-binary-valued by analyzing the specific options selected on multiple-choice questions. Another exception is \cite{okt}, which uses large language models to predict open-ended student responses in a generative way. To represent questions and skills, most existing KT methods one-hot encode them based on question IDs or skill tags \cite{qdkt}, except \cite{okt,eernna}. To represent time steps, most existing KT methods treat each question as a discrete time step, with a few exceptions such as \cite{wang2021temporal}, which considers the exact, continuous time elapsed between responses.

\subsection{Causal Analysis Methods}

In the field of education, there exist very few works on causality and especially few in the context of KT. 
\cite{kaur2019causal} is closely related to our work, where the authors study the relationship between courses in higher educational institutions using historical student performance data. They use matching methods and regression to determine the average treatment effect (ATE). Along similar lines, \cite{sales2018using} and \cite{sales2021student} developed theory and methods for analyzing A/B testing data and presented studied data collected from real-world randomized controlled trials. These works focus on causal inference, i.e., assuming that the structure is given and the focus is on estimating the treatment effect. However, the data we use from Eedi contains fine-grained skills, defined as the smallest elements of learning among primary/middle school students. Therefore, our work is different from these works in terms of both the goal and the educational context. The only existing works that study causal discovery in the context of KT are \cite{tracekdd} and \cite{minn2022interpretable}. The former uses a special model structure that has some similarity to ours to model knowledge state transitions among uninterpretable latent skills. The authors showed that their method, while simple, is highly accurate in predicting unobserved student responses, but do not evaluate on whether the identified causal structure is valid. Our proposed method to learn latent causal structure is closely based on the structural equation model (SEM) \cite{pearl}. SEM enables us to estimate the relationships between observed and latent variables, offering valuable insights into their underlying relationships. 
The hypothesized causal relationship among variables is represented as a directed acylic graph (DAG). In this work, our goal is to learn the causal structure graph $\mathcal{G}$. 

\section{Methodology}

We now detail our conceptual causal KT method. 

\subsection{Basic Setup} 
The basic KT model contains two components:
\begin{align}
    \mathbf{h}_{j,t} \sim f(\mathbf{h}_{j,t-1},\mathbf{x}_{j,t}). \label{eq:ke} \\
    p(Y_{j,t} \: | \: \mathbf{h}_{j,t}, {i}_{j,t}) \label{eq:rp}.
\end{align}
For a student $j$ at time step $t$, The knowledge estimation component in Eq.~\eqref{eq:ke} estimates the current knowledge state $\mathbf{h}_{j,t}$ given the previous knowledge state $\mathbf{h}_{j,t-1}$ and the student's performance on the problem $\mathbf{x}_{j,t}$ as inputs. The response prediction component in Eq.~\eqref{eq:rp} outputs the prediction of the student's likelihood of answering the next question ${Y}_{j, t}$ correctly given the current knowledge state $\mathbf{h}_{j,t}$ and the next question index ${i}_{j,t}$ as input. 
During the learning process, the KT model needs to maximize the predicted likelihood across responses of all students, i.e., 
$\sum_{j} \sum_{t} \log p(Y_{j,t} \:|\: Y_{j,1}, \ldots, Y_{j,t-1}).$

We adopt the DKT setup detailed in \cite{piech2015deep} for consistency. Since incorporating causal learning into the base KT model introduces additional parameters, we use the gated recurrent unit (GRU) as the transition model instead of long short-term memory (LSTM) for computational efficiency. For response prediction, we simply use a single linear layer over the hidden states of the GRU. For causal discovery, i.e., learning the causal structure among skills, we use the causal GRU module detailed below.

\begin{figure}
\centering
\includegraphics[width=0.8\linewidth]{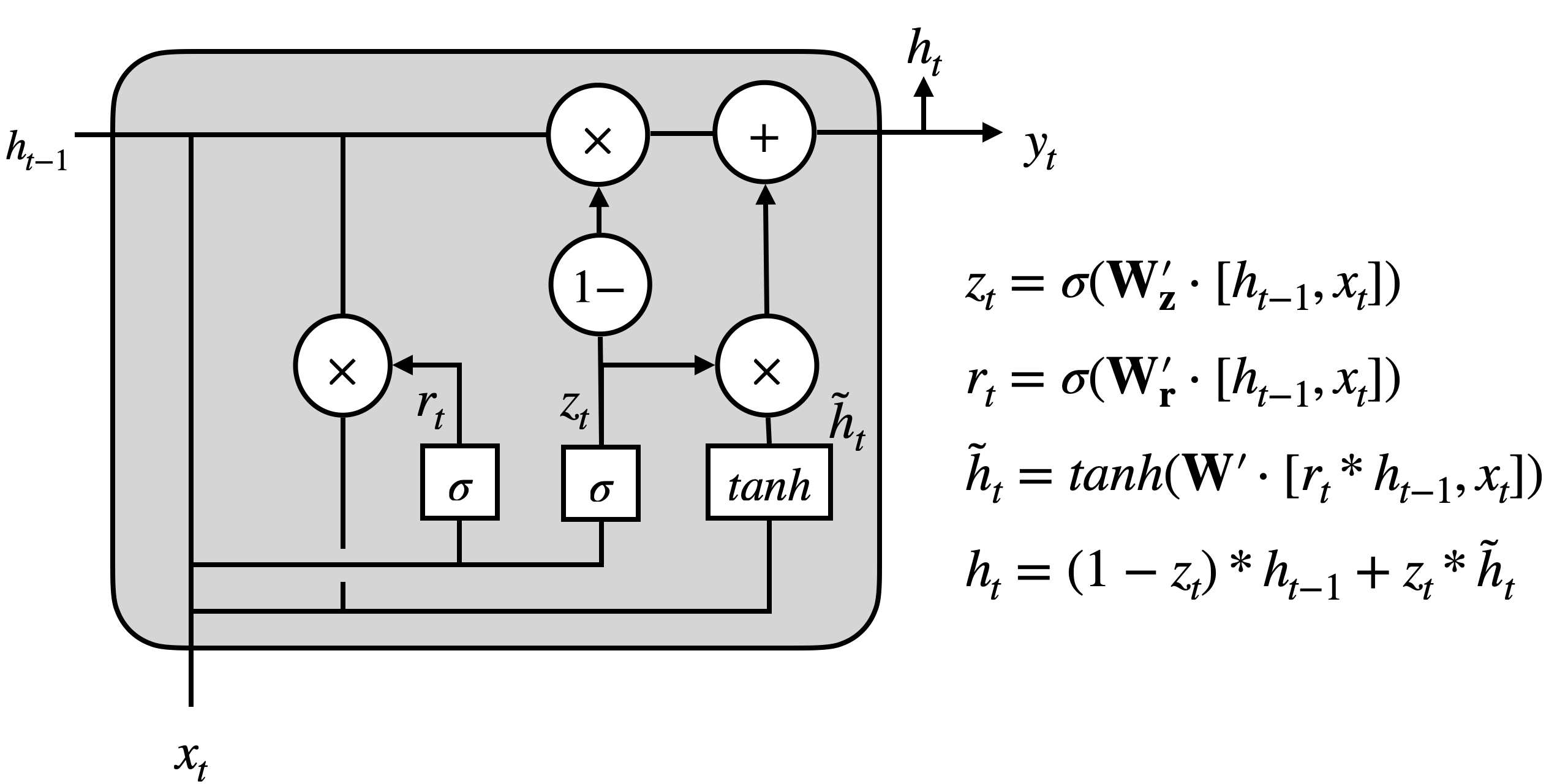}
\caption{The implementation of a causal GRU cell. All the GRU weight matrices, $\mathbf{W}_z$, $\mathbf{W}_r$, and $\mathbf{W}$ are multiplied by the causal mask $\mathbf{M} = \mathbf{P}\mathbf{L}\mathbf{P}^T$, resulting in $\mathbf{W}_{z}'$, $\mathbf{W}_{r}'$, and $\mathbf{W}'$.}
\label{fig:causalgru}
\end{figure}

\subsection{Causal GRU} 
\label{sec:causalgru}

We now detail the structure of the Causal GRU. From now on, we drop the student index $j$ for notation simplicity. We define a permuted causal mask $\mathbf{M}$ that represents the causal ordering and structure between skills. The $\mathbf{L}$ matrix represents the \emph{causal structure/skill dependency}, and the $\mathbf{P}$ matrix represents the \emph{causal ordering}. The permuted causal mask $\mathbf{M}$ is calculated in Eq.~\eqref{eq:tmatrix} as first multiplying $\mathbf{P}$ by $\mathbf{L}$ to obtain the updated causal structure and multiplying by $\mathbf{P}^T$ to transform the causal structure into the original space. 

The parameters in the Causal GRU are masked, i.e., element-wise multiplied by the permuted causal mask $\mathbf{M}$ in Eq.~\eqref{eq:mask}. By masking out some parameters, we zero out parameters that do not satisfy the causal graph. This step ensures that there is no relationship between the hidden states of the non-causally dependent skills in latent student knowledge states. The latest student knowledge state estimation $\mathbf{h}_{t}$ is calculated in Eq.~\eqref{eq:cgru}. The input $\mathbf{x}_{t}$ is represented as a one-hot vector with the dimension size equals to the number of skills $C$. The entry of value $\pm 1$ represents whether the student can correctly answer the question corresponding to the skill. The input $\mathbf{h}_{t-1}$ is the previous student knowledge state estimation. The implementation detail of a Causal GRU cell can be found in Fig.~\ref{fig:causalgru}.

\begin{align}
\mathbf{M} = \mathbf{P} \mathbf{L} \mathbf{P}^T, \label{eq:tmatrix}\\
\mathbf{W}^{'} = \mathbf{M} \odot \mathbf{W}, \label{eq:mask} \\
\mathbf{h}_t = GRU_c(\mathbf{h}_{t-1},\mathbf{x}_t).  \label{eq:cgru}
\end{align}

\begin{figure}[t]
\centering
\includegraphics[scale=0.13]{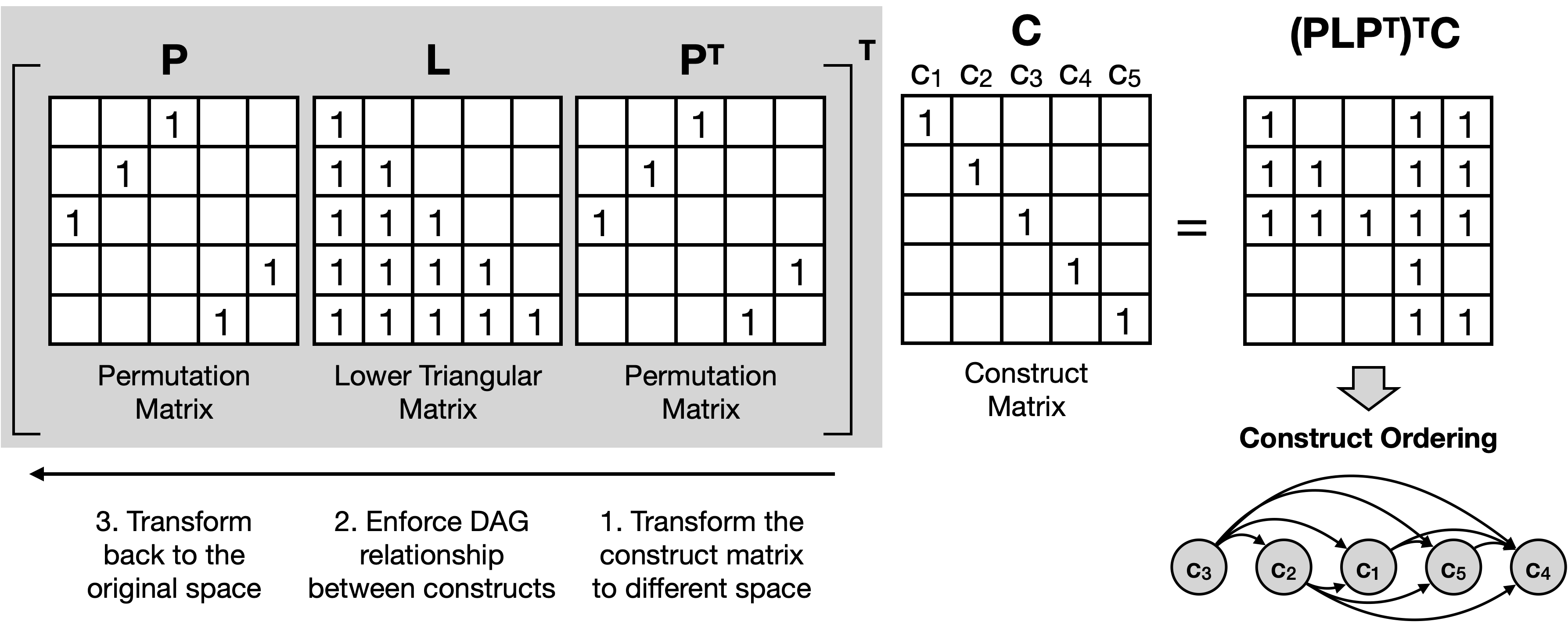}
\caption{The intuition behind causal GRU. Here, $\mathbf{M} = \mathbf{P}\mathbf{L}\mathbf{P}^T$. $M_{i,j} = 1$ if and only if $[\mathbf{h}_{t-1}]_j$ influences $[\mathbf{h}_{t}]_i$ where $\mathbf{h}_t$ is the student's knowledge state at time-step $t$.}
\label{fig:causalkt}
\end{figure}

\subsubsection{Causal Ordering}

One important element of the causal GRU is the \emph{causal ordering} matrix $\mathbf{P}$, which we set to be a permutation matrix. By definition, a permutation matrix has exactly one entry of 1 in each row and each column and 0s elsewhere. Since multiplying a matrix by a permutation matrix permutes the order of the columns/rows of that matrix, the permutation matrix is naturally capable of sorting skills into order based on prerequisite relationships. However, the binary and discrete nature of the permutation matrix makes the learning process non-differentiable. To solve this problem, we introduce a relaxed version of the problem by approximating a permutation matrix with a doubly stochastic matrix, i.e., one where all entries are non-negative and the summation of each column/row is equal to 1, i.e., $ P_{i,k} \in [0,1], \:\sum_{k} P_{i,k} = 1 \: \forall i \: , \: \sum_{i} P_{i,k} = 1 \: \forall k \:$. 
Instead of learning a doubly stochastic matrix directly, which is very difficult, we learn a matrix of free parameters $\bar{\mathbf{P}}$, from which we can obtain $\mathbf{P}$ after applying the \emph{Sinkhorn} operator \cite{sinkhorn1964relationship} $\mathbf{P} = \text{Sinkhorn}(\bar{\mathbf{P}})$.

The \emph{Sinkhorn} operator works as follows: First, starting with the base matrix $\bar{\mathbf{P}}$, we subtract the largest entry of the matrix from each entry, multiply each entry with the temperature hyper-parameter, and pass it through an exponential function. Second, we apply a series of row and column normalizations by dividing each entry of the column/row by the summation of all the entries in the column/row. In our implementation of the \emph{Sinkhorn} operator, there are two hyper-parameters: \emph{temperature} and \emph{unroll}. The temperature hyper-parameter specifies the extent of the continuous relaxation: the larger the temperature hyper-parameter, the closer $\mathbf{P}$'s entries are to either $0$ or $1$. The unroll hyper-parameter specifies the number of times row/column normalization is carried out: the more times the normalization is applied, the closer $\mathbf{P}$ is to satisfying the row/column normalization constraints. 

\subsubsection{Causal Structure} 
The other important element of the causal GRU is the \emph{causal structure/ skill dependency} matrix $\mathbf{L}$, which we set to be lower triangular. By definition, a lower triangular matrix is one in which all the elements above the principal diagonal of the matrix are 0. 
This matrix is important since it specifies the causal structure among the skills. Once the skills are ordered using the \emph{causal ordering} matrix $\mathbf{P}$, we apply the \emph{causal structure} matrix $\mathbf{L}$ to regularize student knowledge state transitions across time steps. Due to its lower-diagonal structure, an entry $L_{i,k} > 0$ with $i>k$ implies that skill $k$ is a prerequisite of skill $i$. Therefore, since the causal GRU weight matrices are masked by the $\mathbf{PLP^T}$ matrix, at the next time step $t$, the entry in the latent student knowledge vector that corresponds to skill $i$, $[\mathbf{h}_t]_i$, depends only on the entries in the previous knowledge state that correspond to prerequisites of skill $i$, i.e., $[\mathbf{h}_{t-1}]_k\,\forall k$ s.t.\ $L_{i,k} > 0$. 

The L matrix being lower diagonal ensures that the resulting causal structure is a DAG. This means that if skill $C_1$ depends on $C_2$ then $C_2$ cannot depend on $C_1$. As a concrete example, Fig.~\ref{fig:causalkt} visualizes the effect of applying a mask $\mathbf{M} = \mathbf{P}\mathbf{L}\mathbf{P}^T$ on the skill matrix $C$. The skill matrix $C$ represents 5 skills where each skill is represented as a one-hot vector. The causal ordering matrix $\mathbf{P}$ is applied to the skill matrix to give a skill ordering of $C_3$, $C_2$, $C_1$, $C_5$, $C_4$ which is in the order of decreasing pre-requisites ($C_3$ being the most pre-requisite of all skills). We further apply the $\mathbf{L}$ matrix (in this case a lower diagonal matrix with all ones) that specifies that every subsequent skill depends on the preceding one. For example, it specifies that $C_4$ depends on all of $C_1$, $C_2$, $C_3$, and $C_5$; $C_5$ depends on $C_1$, $C_2$, and $C_3$ and so on. 

One easy choice for $\mathbf{L}$ is to set its lower-diagonal part to be all ones; this setting means that every subsequent skill causally depends on all the previous skills. However, in practice, causal dependencies among skills nay not be this dense; most skills will only be causally related to a few other skills. To resolve this problem, we can make the $\mathbf{L}$ matrix learnable by restricting the lower diagonal elements to be either 0 or 1. We do this by learning a matrix of free parameters $\bar{\mathbf{L}}$, from which we can obtain $\mathbf{L}$ after applying the element-wise sigmoid operator $\mathbf{L} = \text{sigmoid}(\alpha \bar{\mathbf{L}})$. A large value of the temperature parameter $\alpha > 0$ will push entries to be close to either 0 or 1 but not in between. 


\subsubsection{Skill Embeddings}
\label{sec:modelvariants}

We use a learnable dense embedding to represent each skill and alter both the input and the output layers of the causal GRU. We learn an embedding matrix $\mathbf{E}$ where each column $\mathbf{e}_c$ represents the embedding of skill $c$. We treat the dimension of $\mathbf{e}_c$ as a hyperparameter. For the input layer, we use another learnable embedding $\mathbf{d}$, which is either added or subtracted from the skill embedding depending on the correctness of the previous answer. We then learn the input to the causal GRU using $NN(\mathbf{e}_c \pm \mathbf{d})$ where $NN$ is a single-layer neural network. For the output, we use $p(Y_t) \sim NN_{o}([\mathbf{e}_c^T,\tilde{\mathbf{h}}_t^T]^T)$, where $\tilde{\mathbf{h}}_t$ is a masked version of $\mathbf{h}_t$ with the only non-zero entry being the one that corresponds to the skill of the next question that we are predicting. Here $NN_{o}$ is a single-layer neural network that predicts the probability of the correct answer.

\section{Experiments}

\subsection{Data and Challenge Description} 

We participated in Task 3 of the NeurIPS Challenge co-hosted by Eedi, Microsoft Research, and Rice University \cite{gong2022instructions}. The goal of this task is to discover the causal relationships between different skills, or \emph{constructs} (as defined by Eedi, which means the smallest unit of learning; for example, ``mental addition and subtraction'' is a construct within the main topic ``math''), and evaluate the effect of learning one \emph{skill} on another. 
Questions in this dataset are multiple-choice, with a single correct option and three distractors that are designed to assess a single skill. The challenge hypothesis is that it is possible to discover the hidden relationship behind different skills through analyzing the responses to a large number of diagnostic questions. The challenge uses an $F_1$ score-based metric which calculates the similarity between the predicted adjacency matrix $\hat{A}$ and the true adjacency matrix $A$.

\begin{table}[t]
\centering
\small
\caption{Results on different model variants.}
\begin{tabular}{|l|l|} 
\hline
\begin{tabular}[c]{@{}l@{}}\textbf{Model }\\\textbf{Variant}\end{tabular}        & \begin{tabular}[c]{@{}l@{}}\textbf{Leaderboard}\\\textbf{$F_{1}$ Score}\end{tabular}  \\ 
\hline
\begin{tabular}[c]{@{}l@{}}No Embedding\\No Adaptive\end{tabular}                & 0.11                                                                             \\ 
\hline
\begin{tabular}[c]{@{}l@{}}No Embedding\\Adapative\end{tabular}                  & 0.17                                                                             \\ 
\hline
\begin{tabular}[c]{@{}l@{}}Embedding (300D)\\Adaptive\end{tabular}               & 0.33                                                                             \\ 
\hline
\begin{tabular}[c]{@{}l@{}}Embedding (300D)\\Adaptive~\\Learnable L\end{tabular} & 0.43                                                                             \\
\hline
\end{tabular}
\label{tab:results1}
\end{table}

\subsection{Model Learning and Hyperparameters}

The dataset consists of 1855 skills and 6468 students.  We set the default skill embedding dimension to 300. We use an adaptive strategy and start with small values of the temperature and unroll and linearly increase their values over a set of epochs. We set the initial temperature and unroll to 2 and 5 respectively and linearly increase the values with a factor of 2 and 5 respectively for every 10 epochs. We train the model for 50 epochs with a batch size of 64, and a learning rate of 5e-4 using four Nvidia Tesla 2080 GPUs with a GPU memory of 12GB each which takes about 6 hours. After training, to obtain the final \emph{causal structure} matrix $\mathbf{L}$ we apply a post-processing step. We define a hyperparameter \emph{$\kappa$} such that all values of the $\mathbf{L}$ matrix less than \emph{$\kappa$} are set to 0 and all values greater than or equal to \emph{$\kappa$} are set to 1.

\subsection{Results and Discussion}

In Table~\ref{tab:results1}, we show the results of different model variants. We report the leaderboard $F_1$ score obtained in our experiments. We see that the $F_{1}$ score is 0.11 for the case where we are not using the skill embeddings. 
Using an adaptive strategy increases the $F_{1}$ score by 0.06, which suggests that the adaptive strategy is helpful during model training. We also report the results corresponding to the skill embeddings and the learnable $\mathbf{L}$. We see that using an embedding dimension of 300 almost doubles the $F_{1}$ score. This observation confirms our hypothesis that using skill embeddings increases the representational capacity of the neural network model and hence performs better. When the causal structure matrix $\mathbf{L}$ is learnable, we see that we get a further 0.1 increase in the $F_{1}$ score. The increase in the $F_{1}$ score on using the learnable $\mathbf{L}$ configuration of the model shows that it is better to learn the explicit causal dependence of skills instead of assuming a dense representation where each skill depends on all the skills preceding it.

\section{Conclusions and Future Work}

In this work, we proposed a conceptual method for learning causal structure among skills from student response data, as a part of our solution to the NeurIPS 2022 Challenge on Causal Insights for Learning Paths in Education. Our method is a novel causal knowledge tracing method that enables us to learn the causal structure in an end-to-end manner while performing knowledge tracing. Unfortunately, due to space limitations, we cannot show a qualitative example of the learned causal structure among skills. We believe that our work should inspire future works in the direction of building causal knowledge tracing methods on observational student response data. 
First, it is important to evaluate the accuracy of the learned causal structure between skills, either against human domain experts or via A/B testing. 
Second, it is important to apply our causal module to more flexible knowledge tracing methods, such as attention-based methods, to see whether it is applicable and effective. 
Third, it is important to develop ways to leverage both the opinion of human experts and our data-driven causal discovery model, in a human-in-the-loop manner. The former may be less accurate but the latter requires extensive training data; a hybrid human-AI collaboration may be able to take the best from both sides. 

\section{Acknowledgements}
The authors thank Christoph Studer for insights on permutation matrices and the NSF (under grants 1917713, 2118706, 2202506, 2215193) for partially supporting this work. We also thank the organizers of the 2022 NeurIPS Challenge on Causal Insights for Learning Paths in Education for proposing a new and meaningful task. 

\clearpage

\bibliographystyle{abbrv}
\bibliography{main}

\clearpage

\appendix

\section{Structural Equation Modeling} 

In SEM, given a collection of random variables $\mathbf{x} = (x_1, \cdots, x_D)$ and a causal directed acyclic graph (DAG) $\mathcal{G}$, each variable $x_i$ is generated using its parents $Pa(i;\mathcal{G})$ in the DAG and an exogenous noise variable $\epsilon_i$:  $x_i = f_i(\{x_j\}_{j\in Pa(i;\mathcal{G})}) +\epsilon_i$. The structural vector autoregression (SVAR) model extends SEM to random variables that form a time series $\mathbf{x}_t = (x_{1,t}, \cdots, x_{D,t})$ for time steps $t\in \{1,\cdots,T\}$ \cite{varlingam}. The influence of one variable on others can be instantaneous or lagged behind for a few time steps. The SEM model is given by
\begin{align*}
    x_{i,t} =  \sum_{\tau=0}^k f_{ i,\tau}(\{x_{j, t-\tau}\}_{j \in Pa(i, t,\tau;\mathcal{G})}) +\epsilon_{i,t},
\end{align*}
where ${Pa(i, t,\tau;\mathcal{G})}$ are random variables from time step $t-\tau$ that influence random variable $x_{i,t}$. 
One can also use a single latent state $\mathbf{h}{\cdot,t}$ to model the influence of past random variables. The SEM becomes
\begin{align*}
    \mathbf{h}_{i,t} =  f_{i} (\{h_{j,t-1}\}_{j\in Pa(i;\mathcal{G})})+\epsilon_i.
\end{align*}

\section{Sinkhorn Operation}

\begin{figure}[t]
\centering
\includegraphics[width=0.4\linewidth]{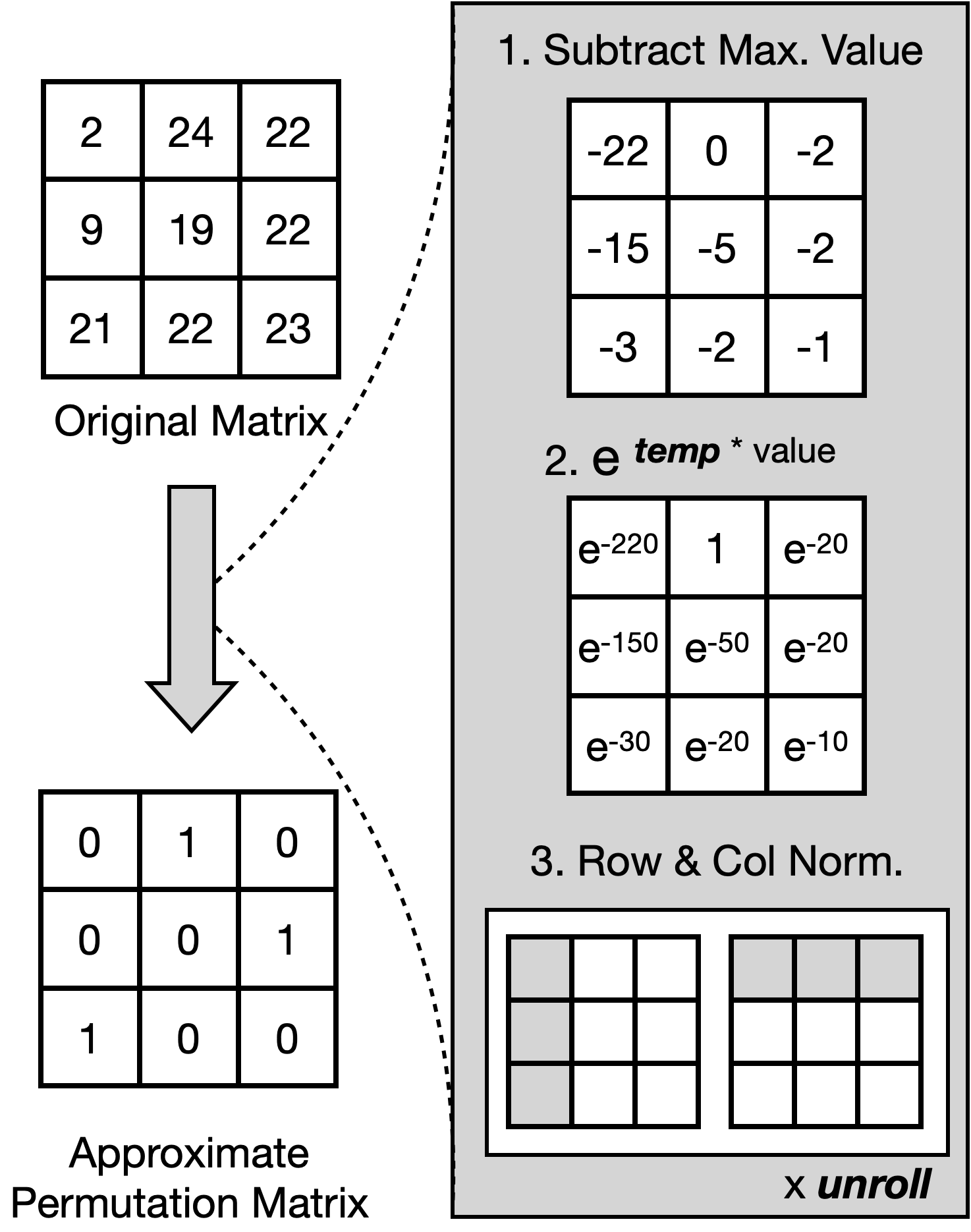}
\caption{The \emph{Sinkhorn} operator is a smooth operator that outputs an approximate permutation matrix.} 
\label{fig:sinkhorn}
\end{figure}

Fig.~\ref{fig:sinkhorn} shows the working of the Sinkhorn operator. 

\section{Additional Results}

\begin{table}[t]
\centering
\caption{Results on varying the cutoff for the $\mathbf{L}$ matrix, $\kappa$.}
\begin{tabular}{|l|l|} 
\hline
\textbf{$\kappa$}   & $F_1$ Score  \\ 
\hline
0.42  & 0.43      \\ 
\hline
0.45  & 0.43      \\ 
\hline
0.435 & 0.43      \\ 
\hline
0.48  & 0.43      \\ 
\hline
0.495 & 0.43      \\ 
\hline
0.51  & 0.33      \\ 
\hline
0.525 & 0.18      \\
\hline
\end{tabular}
\label{tab:tauvarying}
\end{table}

We report the results obtained using different hyperparameters of the learnable $\mathbf{L}$ model configuration. In Table~\ref{tab:tauvarying}, we show the results across different \emph{$\kappa$} values. We vary the values from 0.42 to 0.525 and observe that using any values out of this range gives a leaderboard $F_{1}$ score of 0. Among the experimented values for \emph{$\kappa$} we see that we obtain a maximum $F_{1}$ score of 0.43 for all values in the range of 0.42 to 0.495. The maximum $F_{1}$ score for \emph{$\kappa$} values in the range of 0.42 to 0.495 means that using very large or very less values of \emph{$\kappa$} does not give the optimal skill dependency.

\begin{figure}[t]
\centering 
\includegraphics[width=\linewidth]{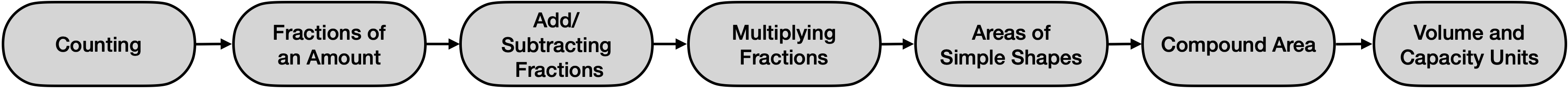}
\caption{An example of learned causal ordering among skills in actual student response data.}
\label{fig:dagexample}
\end{figure}

We perform a qualitative analysis of our proposed method. Fig.~\ref{fig:dagexample} shows an example from the DAG obtained from the learned adjacency matrix for causal relations. Here, we represent the constructs based on their subject names, and an arrow from subject $i$ to subject $j$ implies that subject $i$ is a pre-requisite of subject $j$. Here, in the figure, we can see that ``Counting'' is the most pre-requisite skill. The subsequent skills in fractions depend on ``Counting''. Calculating areas of simple figures depends on fraction multiplication. In the same way, calculating the volume depends on calculating the area. Hence, from this, we can see that we are able to learn a meaningful DAG using our methodology.

\balancecolumns
\end{document}